\documentclass[sigconf]{acmart}

\usepackage{color,soul}
\usepackage{subfig}
\usepackage{subfloat}
\usepackage{multirow}
\usepackage{algorithm, algorithmic}

\AtBeginDocument{%
  \providecommand\BibTeX{{%
    \normalfont B\kern-0.5em{\scshape i\kern-0.25em b}\kern-0.8em\TeX}}}

\setcopyright{acmcopyright}
\copyrightyear{2022}
\acmYear{2022}
\acmDOI{XXXXXXX.XXXXXXX}

\acmConference[Conference acronym 'XX]{Make sure to enter the correct
  conference title from your rights confirmation emai}{June 03--05,
  2022}{Woodstock, NY}
\acmPrice{15.00}
\acmISBN{978-1-4503-XXXX-X/18/06}

\begin{document}

\title{Split Federated Learning on Micro-controllers: A Keyword Spotting Showcase}


\author{Jingtao Li}
\affiliation{%
  \institution{ASU}
  \city{Tempe}
  \country{USA}}

\author{Runcong Kuang}
\affiliation{%
  \institution{ASU}
  \city{Tempe}
  \country{USA}
}






\renewcommand{\shortauthors}{Jingtao and Runcong, et al.}

\begin{abstract}
  Nowadays, AI companies improve service quality by aggressively collecting users' data generated by edge devices, which jeopardizes data privacy.
  To prevent this, Federated Learning is proposed as a private learning scheme, using which users can locally train the model without collecting users' raw data to servers. 
  However, for machine-learning applications on edge devices that have hard memory constraints, implementing a large model using FL is infeasible.
  To meet the memory requirement, a recent collaborative learning scheme named split federal learning is a potential solution since it keeps a small model on the device and keeps the rest of the  model on the server. In this work, we implement a simply SFL framework on the Arduino board and verify its correctness on the Chinese digits audio dataset for keyword spotting application with over 90\% accuracy. Furthermore, on the English digits audio dataset, our SFL implementation achieves 13.89\% higher accuracy compared to a state-of-the-art FL implementation.

\end{abstract}





\maketitle

\section{Introduction}

Increasing accessibility of edge computing devices has shown great impact on our daily life, and at the same time, their generated data powers up machine learning applications such as computer vision and voice recognition. However, as these devices are close to our living environment, it starts to raise concern on the data privacy. General Data Protection Regulation (GDPR) that is recently approved by European Union imposes a strong limitation on collecting data from users. In this context, how to utilize these valuable data while comply with GDPR is still a challenge for AI community. 

In fact, private schemes such as multi-party computation (MPC)\cite{goldreich1998secure} and homomorphic encryption (HE) can enable provable privacy while handling users' data. However, the large overhead of computation and communication of these private methods are not practical for public use (though they can be good for cross-silo use case). Thus, companies start to investigate cheap but privacy-preserving learning schemes. Google presents Federated Learning (FL) \cite{mcmahan2017communication} as one of these schemes and deploy it in its Google assistant and Google Keyboard applications.

The key idea of FL is to prevent users' data from leaving the device. In its design, each user acquires a local copy of the model, train it on user's own private data, and send the updated parameters to the server periodically. Server, once collecting updated parameters from all users, perform an average (as in FedAvg\cite{mcmahan2017communication}) to get the global model updated and transfer to user for training in the next round. For the first time, \cite{llisterri2022device} brings the Federated Learning idea to the TinyML community, where they are able to demonstrate the successful implementation of FL in a tiny micro-controller for a 2-layer MLP model in a binary keyword classification problem.

A clear limitation of using FL is that it requires user have the ability to train the entire model locally on the device. For embedded systems which are extremely constrained by memory and power, training locally is not possible for an industry-scale model. Thus, \cite{llisterri2022device} uses a tiny 2-layer MLP of 16K parameter, mostly in the concern of the limited SRAM of the micro-controller.

For more complex datasets and use cases, model has to be larger. Thus, we investigate the realization of Split Federated Learning (SFL) \cite{thapa2022splitfed} as an edge-friendly version of FL on a micro-controller. SFL, splits the model into two parts, client-side model and server-side model. Client-side model is designed to be small enough to run on the device and server-side model can be very large to achieve a good accuracy.
As a split version of Federated Learning, the raw data is also protected in SFL since it is processed pure in local devices. Moreover, SFL allows users to have only a small part of the model which can relieve the hardware requirement.

In the scope of our project, we implemented SFL on Arduino Nano BLE 33 sense for a multi-class keyword classification task, where an audio dataset of digits pronounced in Chinese and English is classified respectively. By SFL, We are able to use very large model in training because server does not have memory limitation and has fast training speed. We compared different model sizes (FL's and SFL's) in terms of the model's classification accuracy, and gave a proof-of-concept example of using SFL can achieve better performance than using FL in practice. We also demonstrated that SFL enables using bigger model to enhance the accuracy. SFL shows 13.89\% better accuracy using a large CNN model compared to Federated Learning where only a small MLP can be used. After training, we compressed the entire model and put it back to the device. Then, the device can perform inference without dependence on the server. Moreover, we realized multi-client training to speed up the process, although they are processing in serial mode. 

In summary, we made several \textbf{contributions}:

\begin{itemize}
    \item For the first time, SFL is deployed on a micro-controller on an extremely low-power and low cost embedded system, for a meaningful digit-spotting application.
    \item We demonstrate using SFL breaks the hard memory limitation of training a large model. The resulting large model can achieve better performance than a small model which is constrained under the case of FL implementation.
    \item As the inference of conventional SFL is also dependent on remote communication with a server, to speed it up, after training is done, we compress the entire model and convert it to a TensorFlow-lite model for easy deployment.
\end{itemize} 

The structure of the rest of the paper is as follows: in section 2 we will go over the related works, system design is introduced in section 3, and a thorough evaluation is done in section 4. Finally, section 5 concludes this work.





\section{Related Work}

\subsection{Secure Computation \& Federated Learning}
Secure multi-party computation and homomorphic encryption are two well-established computational scheme that guarantees the data privacy. However, these are hard to implement for public use and thus heuristic private learning scheme such as FL are proposed and used in practice for cross-client collaborative learning.
FL, first proposed by \cite{mcmahan2017communication}, has achieved great amount of attention. But due to its requirement on user that training needs to be done locally, it may not be the optimal solution for micro-controller use cases, especially to solve meaningful and complicated tasks.

\subsection{Split Federated Learning}
SFL \cite{thapa2022splitfed} has been proposed as a parallel version of Split Learning \cite{vepakomma2018split} (SL), by incorporating the model aggregation idea of FL. It separates the global model into two part, client-side model and server-side model.
At each round, each client gets the up-to-date client-side model from the server, and perform forward propagation on its private data, send only the label of the data, and output (intermediate activation) of client-side model to the server. Server accepts the activation, continue forward propagation on the server-side model and compute the loss on the logits and label. Backward propagation is done at the server and the server-side model is updated, then, the gradients w.r.t. the activation is transmitted to the client from the server. Clients continue the backward-propagation and update its client-model. The above process can happen in parallel for multiple clients and clients can get different copies of client-side model since they train on different data. Thus, at the end of each round, server would collect all the local copies of the client-side model and perform a central aggregation, similar to FedAvg.

Apart from its strength, SFL has two known problems, one is its communication overhead and the other is its vulnerability to model inversion attacks. Recent works \cite{chen2021communication, titcombe2021practical} have addressed these weaknesses in some extent, but these are still need to be cautiously evaluated.

\section{System Design}

\subsection{Overall system}
\textbf{Platform.}
We use Arduino Nano 33 BLE board as the main demonstration platform for this work. The board has a 2-bit ARM Cortex-M4 CPU running at 64 MHz with 256KB SRAM and 1MB Flash memory, operating at 3.3V.

\textbf{Preparation.}
To perform the SFL training, as the data labeling and storage function has not been supported yet. We simulate the data collection, we store the client's data on another PC and transfer to the board using serial port.
Then, the board locally performs the Mel-frequency cepstral coefficients (MFCC) for audio data preprocessing following and pass the MFCC feature to the client-side model.
We use the pc as the server as well as the main control of the whole process including communication, which can also store public and pre-collected dataset for pre-training and training. In addition, We use pc for simulation of additional boards virtually as we don't have more boards. The board equipped with the microphone can be used to collect real-time data for inference, but it's hard to collect data to do the training because it has limited storage space and it's hard to let users label data only on board synchronously, which is left for future work. 

\begin{figure}
    \centering
    \includegraphics[width=\linewidth]{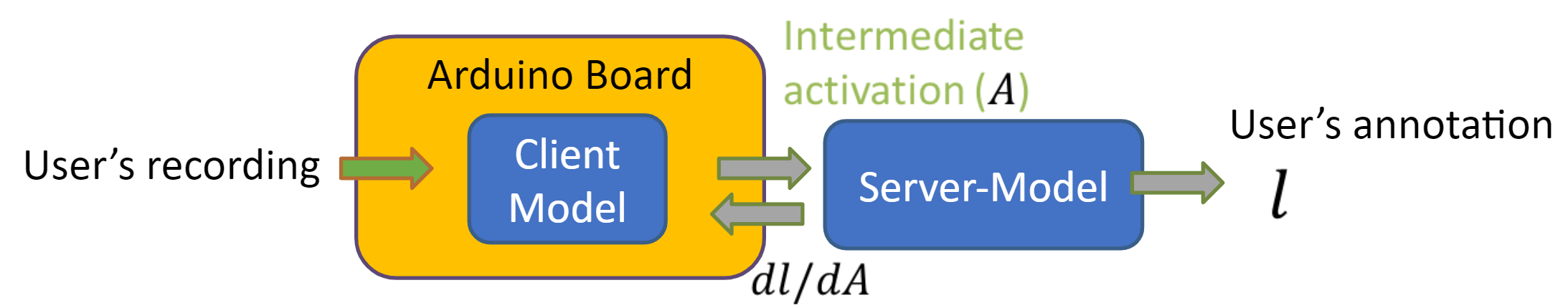}
    \caption{Overall System}
    \label{fig:sys}
\end{figure}

\textbf{SFL training.}
Figure \ref{fig:sys} shows the overall system design of SFL setup. The model is divided into two sides: client model (on-board) and server model (on server/pc). These two parts are connected by the serial link and serial ports. 
The detailed process of SFL training is shown in algorithm~\ref{alg:sfl}. Board preprocesses the raw audio using MFCC and process the MFCC features on client-side model, and send the intermediate activation to the server together with labels. 
Then the server continues the training, using the label and computed logits to generate the loss. Then, server performs backward-propagation and send the gradients to the board.
to get both client-side model and erver-side model updated.
After each epoch of training, local copies of client-side model from multiple devices is aggregated in the server and the aggregated weight is distributed to clients.

\begin{algorithm}
\caption{Split Federated Learning} \label{alg:sfl}
\begin{algorithmic}[1]
\REQUIRE $\quad$ For $M$ clients, instantiate private training data ($X_i, Y_i$) for $1, 2, ..., M$. The entire DNN model of $L$ layers is split to server-side model $S$ and client-side model $C_i$.

\STATE initialize $C_i, S$
\FOR{epoch $t\gets 1$ to num\_epochs}{
    \STATE $C^{*} = \frac{1}{M}\sum_{i=1}^{M}{C_i}$  \hfill\COMMENT{\textcolor{blue}{Model Synchronization}}
    \STATE $C_i \gets C^{*} $ for all $i$
    \FOR{step $s\gets 1$ to num\_batches }{
    \FOR{client $i\gets 1$ to $M$ \textbf{in Parallel}}{
    
    \STATE data batch ($x_i, y_i$) $\gets$ ($X_i, Y_i$)
    \STATE $A_i = C_i(W_{C_i};x_i)$ \hfill\COMMENT{\textcolor{blue}{Client forward; send $A_i$ to Server}}
    }
    \ENDFOR
    \\\hrulefill
    \FOR{client $i\gets 1$ to $M$ \textbf{in Sequential}}{
    \STATE $l = l_{CE}(S(W_{S};A_i), y_i)$ \hfill\COMMENT{\textcolor{blue}{Server forward}}
    
    \STATE $\nabla_{A_i} l  \gets$ back-propagation
    \hfill\COMMENT{\textcolor{blue}{Send $\nabla_{A_i} l$ to Client}}
    }
    \ENDFOR
    
    \STATE Update $W_{S}$;
    \\\hrulefill
    \FOR{client $i\gets 1$ to $M$ \textbf{in Parallel}}{
    \STATE $\nabla_{x_i} l \gets$ back-propagation
    \hfill\COMMENT{\textcolor{blue}{Client backward}}
    \STATE Update $W_{C_i}$;
    }
    \ENDFOR

    }
\ENDFOR
}
\ENDFOR

\end{algorithmic}
\end{algorithm}

\subsection{Data Collection and Preprocessing}

We collected audios of Chinese digits' (from 1 to 5, plus 'silence' and 'unknown') pronunciation from us using the open-source edge impulse platform online. For each class, we collected 50 data. We  We also used corresponding classes from the existed popular \emph{Speech Command}~\cite{warden2018speech} dataset, which has 7000 data in total, in English. We name it as ENdigits. 

For data preprocessing, We used the embedded MFCC in edge impulse to preprocess data first, which is directly deployable on board. We also implement MFCC in python to enable fully training online to speed up the model's hyper-parameters search. For the MFCC hyperparameter, we use a number of coefficients of 13, frame length of 0.02 with 0.02 stride. We use 32 filters and the FFT length of 256. And the normalization window size is set to 101.

To perform the SFL training, as we cannot really collect data and train them in real time, we allow audio input of 1 second (16,000 samples) to be stored on another PC and transfer to the board using serial port.
After the board locally performs the MFCC (as implemented in \cite{llisterri2022device}) followed by audio data preprocessing, MFCC's features are passed to the client-side model.

\begin{figure}[htbp] 
    \centering
    \subfloat[Train Accuracy]{%
        \includegraphics[width=0.49\linewidth]{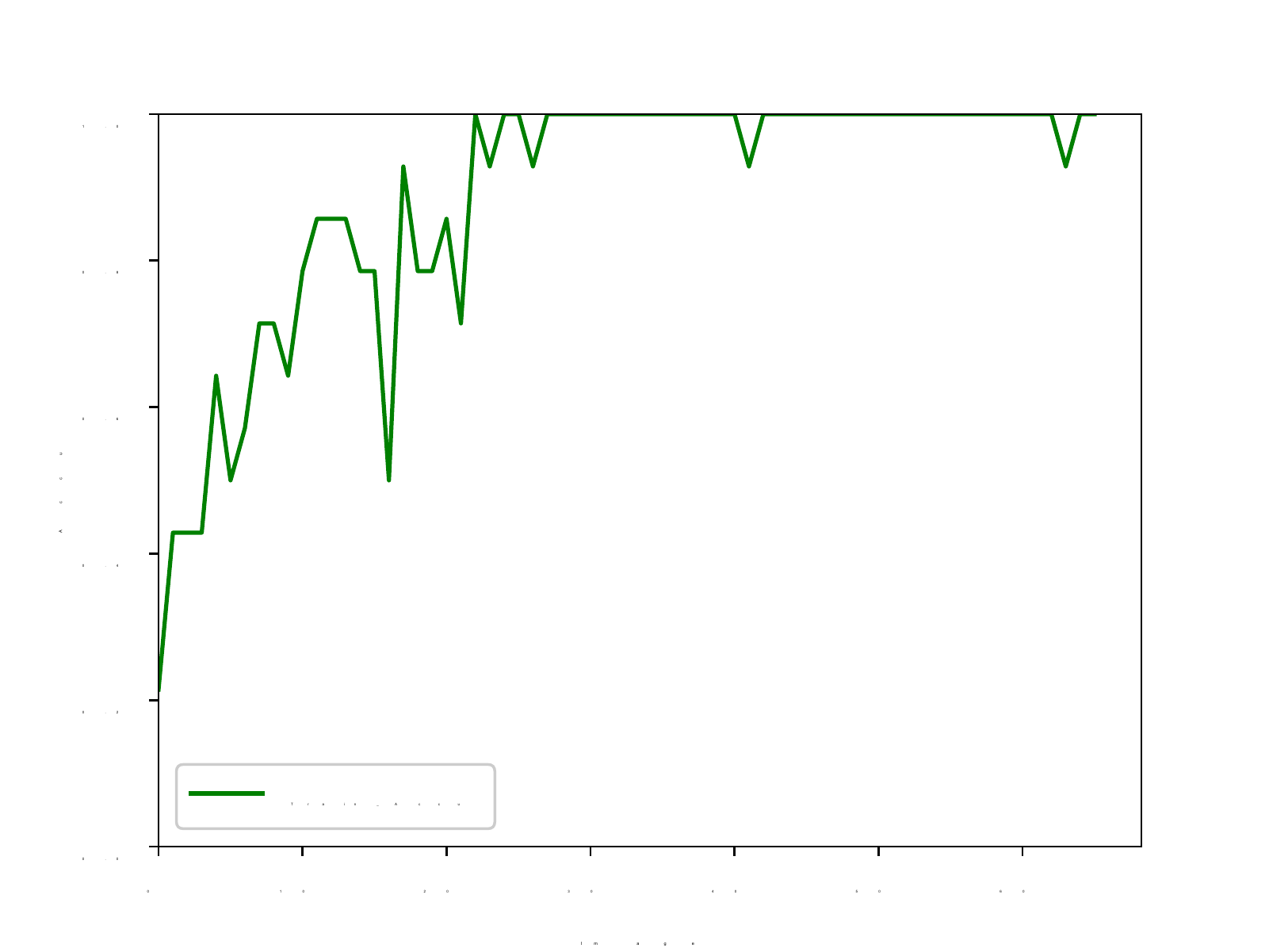}
        \label{fig:mlpsa}%
        }%
    \hfill%
    \subfloat[Validation Accuracy]{%
        \includegraphics[width=0.49\linewidth]{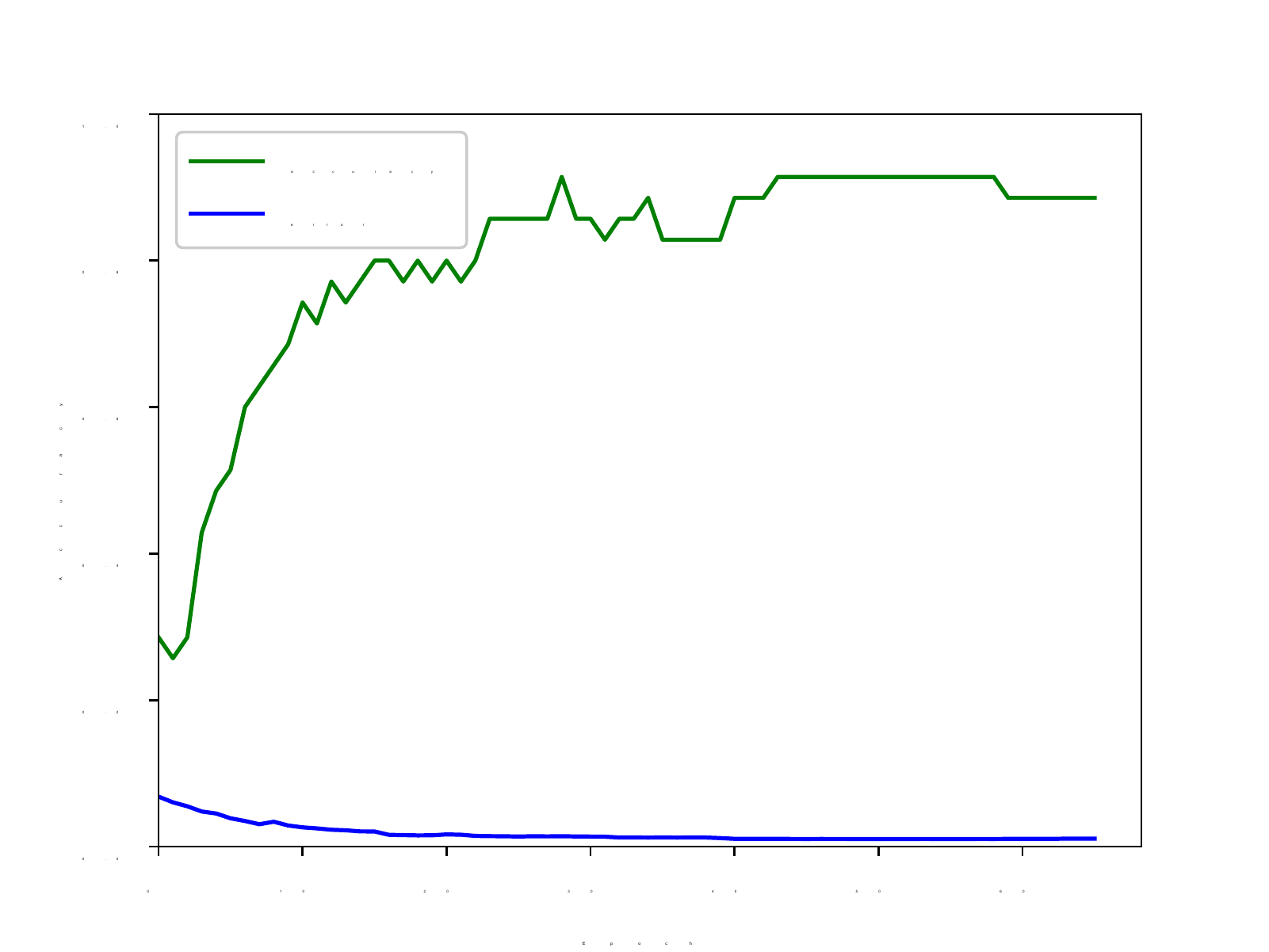}
        \label{fig:mlpsb}%
        }%
    \caption{Split Federated Learning on CNdigits dataset using MLP model.}
    \label{fig:exp_mlp_sfl}%
\end{figure}

\begin{figure*}[htbp] 
    \centering
    \subfloat[Model 1]{%
        \includegraphics[width=0.3\linewidth]{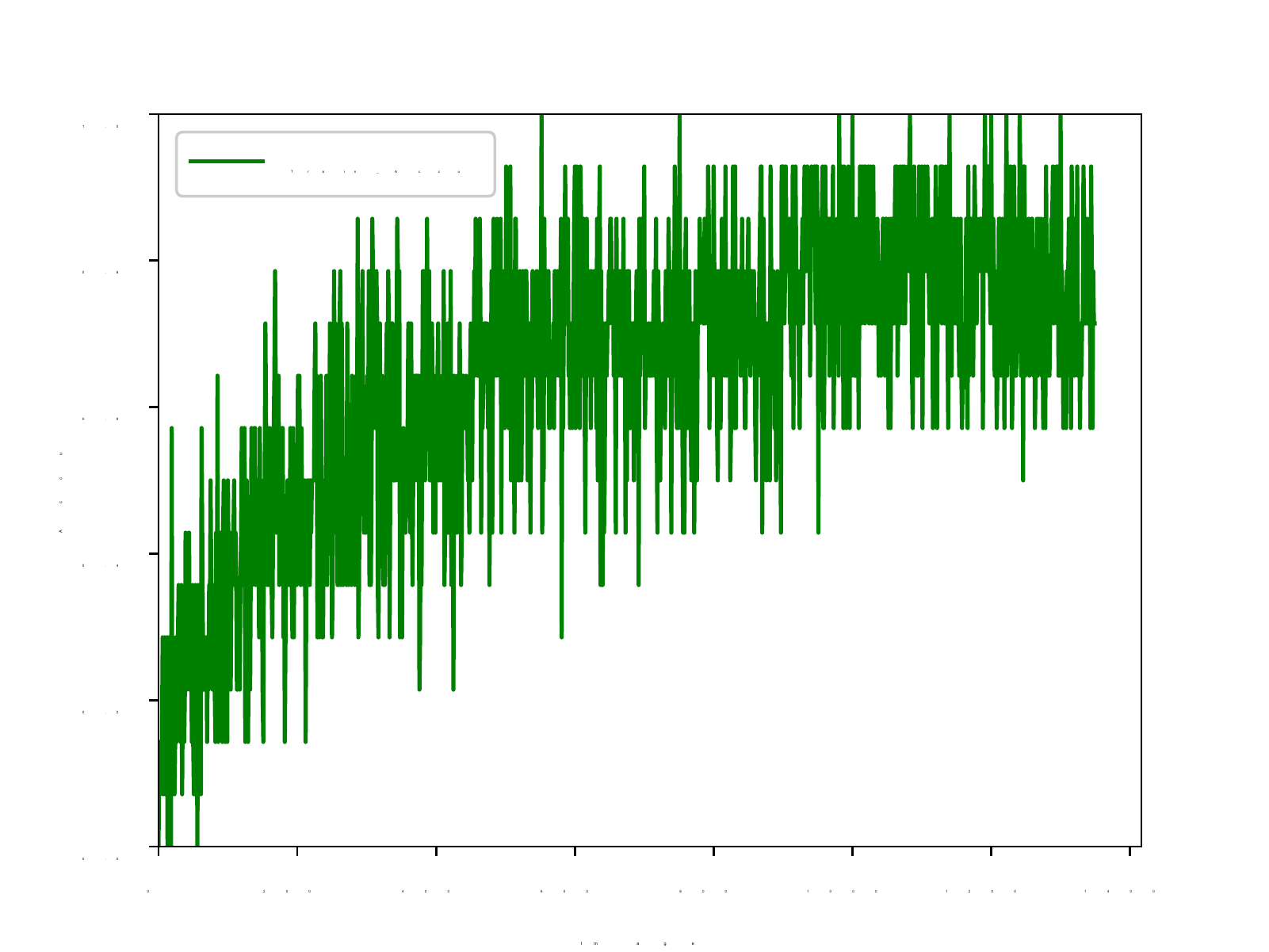}
        \label{fig:train1}%
        }%
    \subfloat[Model 2]{%
        \includegraphics[width=0.3\linewidth]{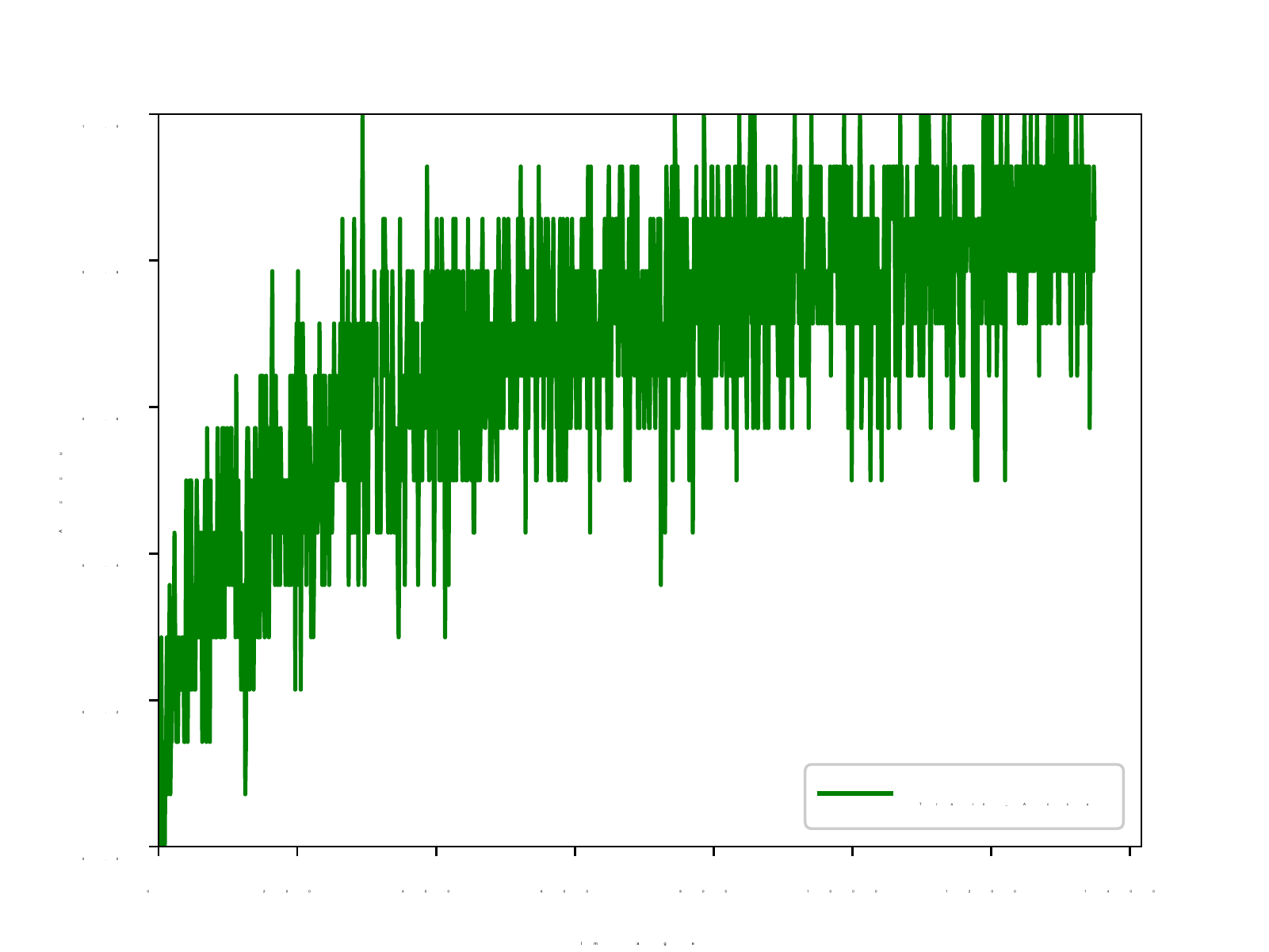}
        \label{fig:train2}%
        }%
    \subfloat[Model 3]{%
        \includegraphics[width=0.3\linewidth]{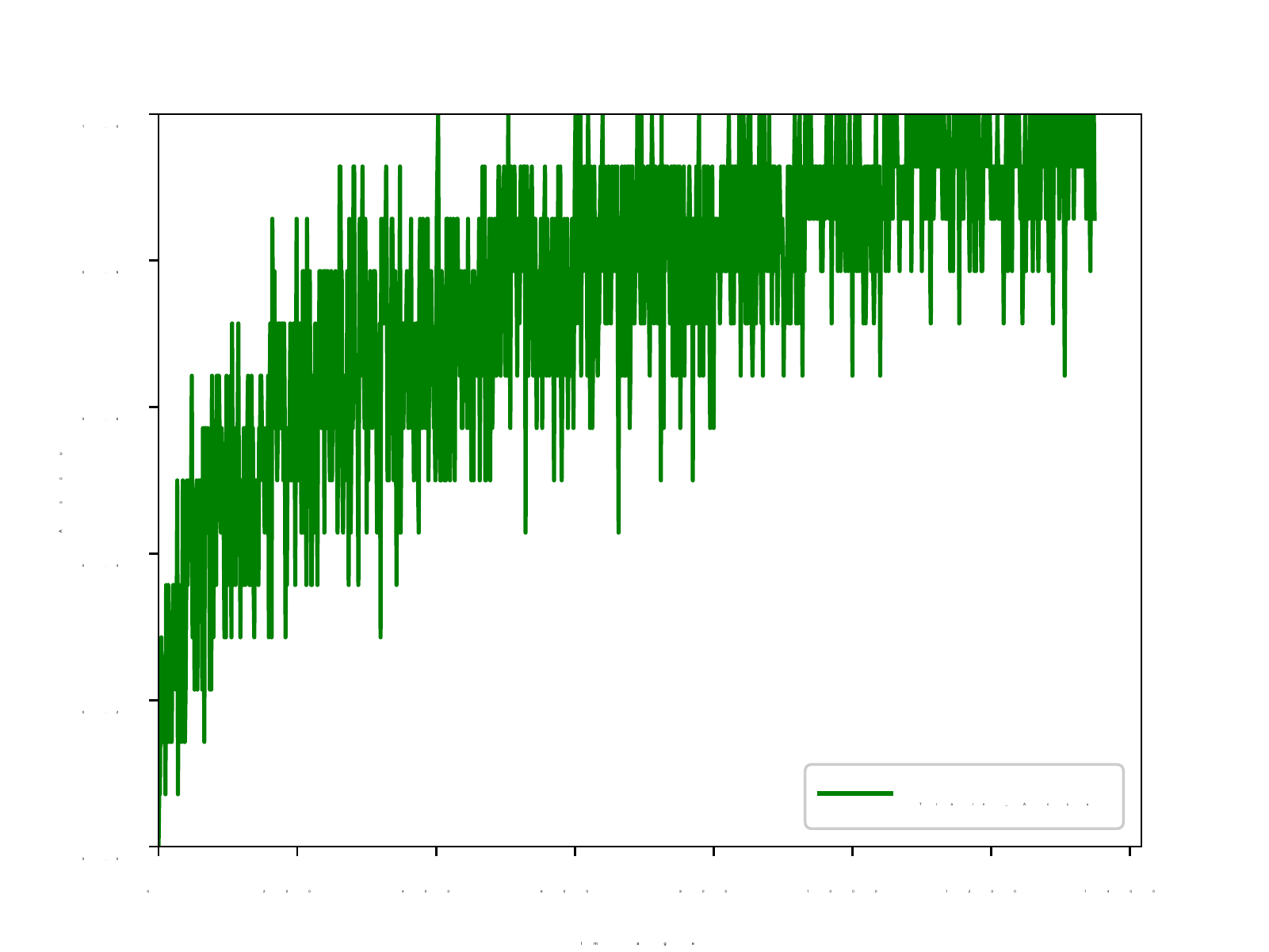}
        \label{fig:train3}%
        }%
    \caption{Training accuracy of 3 models on ENdigits dataset.}
    \label{fig:training}%
\end{figure*}

\begin{figure*}
    \centering
    \includegraphics[width=0.9\linewidth]{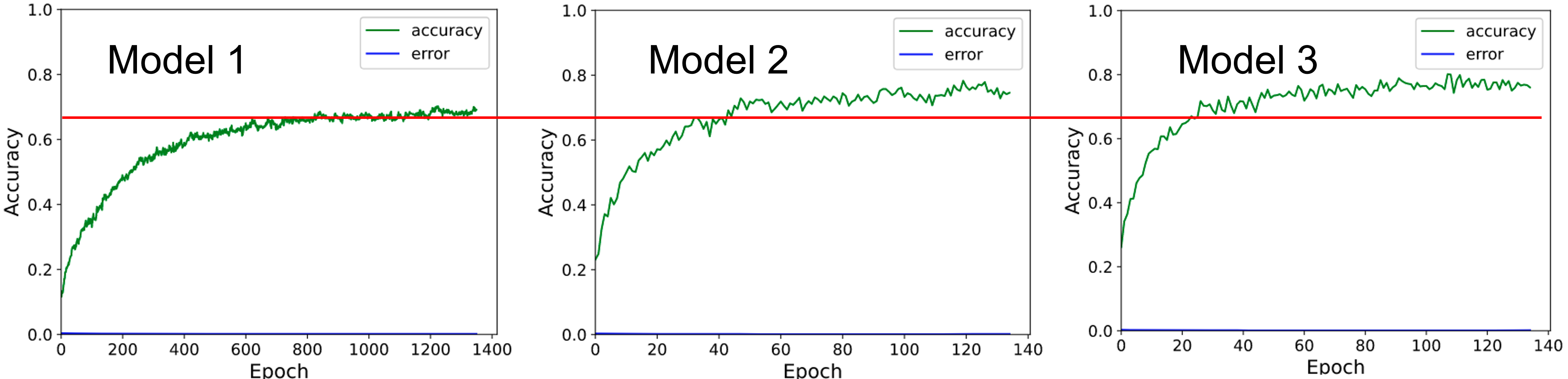}
    \caption{Validation Accuracy of 3 models on ENdigits dataset.}
    \label{fig:validation}
\end{figure*}

\subsection{Model Structure}
Because of the small size of memory space of the board, we need to carefully design the neural network architecture. For the client-side model, given the hardware restriction of the board, we can either choose a small 1-layer MLP model (650x25) on the client model, same as in\cite{llisterri2022device} or a 1-layer Conv2D model with (12x1x3x3) size. 
On the server side, we implement several models with different complexity. The detailed network topology is shown in Table~\ref{tab:network}.

\begin{table}[htbp]
\caption{Network Options}
\centering
\resizebox{1.0\linewidth}{!}{
\begin{tabular}{lcc}
 \toprule
 \textbf{Model} & \textbf{Client} & \textbf{Server}  \\
 \midrule
 Model 1 (MLP)& FC25 & FC7  \\
 Model 2 (CNN)& \multirow{2}{*}{Conv12x3x3} & Conv16x3x3-FC128-FC7   \\
 Model 3 (CNN)&  & Conv30x3x3-FC256-FC7  \\
 \bottomrule
\end{tabular}
}
\label{tab:network}
\end{table}

MLP model is good enough for the easy dataset where feature maps are linearly separable. In the case, higher complexity in server model doesn't help a lot. We choose 2-layer FC (650 * 25 * 7) as our baseline for SFL, which has 650 * 25 + 25 * 7 = 4725 parameters, with 4550 parameters on board.
It turns out that for small CNdigits dataset, MLP would work very well and we don't need a very big CNN model. Typically, a large CNN can cause overfitting problem and gets even worse validation accuracy.
For the larger ENdigits dataset, we observe using a CNN model can get much better performance than the MLP. Thus, we design two Conv2D models for the server-side model, one with small output channel size of 16 and is connected with a FC layer with 128 neurons, and one with larger output channel of 30 and is connected with a FC layer with 256 neurons.

\subsection{Training Hyperparameter}
Due to the limitation of onboard resources, we keep the training batch size as 1. We set learning rate as 0.005 for CNN model and 0.0005 for MLP model. The SGD's momentum is set as 0.6.
The training and validation split ratio is 9:1. The number of epochs for SFL is fixed at 3 epochs.

\subsection{Model implementation}


For training, we used Pytorch to implement the server-side model in the server and use C++ to code the client-side model on the board. We also code the client-side model with MFCC in Pytorch to simulate other clients, to simiulate a multi-client SFL scheme.
After training, we transferred the entire Pytorch model to Keras model, by manually loading the weights, which causes us tons of trouble because of different layout of parameters in these two different framework. After that, the model is converted to TensorFlow lite with int8 quantization.

\section{Evaluation Approach}


We use the validation accuracy (trained model's prediction on the validation dataset) as the criteria to evaluate learning scheme's performance.

\section{Results}

\subsection{Proof of Correctness}

First We prove the correctness of our SFL implementation. We train model 1 using SFL on the Chinese digits (CNdigits) dataset and the result is shown in Figure \ref{fig:exp_mlp_sfl}. We can see the model converges quickly in 70 rounds and converges to a high accuracy of above 90\%.

\subsection{Advantage of SFL}

To show the advantage of SFL, We test both MLP model and CNN model on ENdigits. The FC model represents performance of FL as it cannot train the entire CNN on board. Two CNN models are made possible with SFL.
From the results shown in Figure\ref{fig:validation}, 
all models can reach over 80\% training accuracy after 2 epochs of training. Because of SGD with batch size of 1, accuracy varies a lot in the training period. In addition, CNN shows a faster converge speed.
For validation accuracy, three models converge at 66.25\%, 78.29\%, 80.14\%, respectively.
We conclude that larger and more complex model can generate higher performance than smaller models on ENdigits dataset.

\subsection{Convert to TFlite}
Server collects the final trained global model (combined of the client-side and server-side models). After transformation from Pytorch to Keras, Keras model is transformed into TFlite model with quantization enabled, to better fit into the size of the client device. Finally, the quantized TFlite model is able to be directly deployed on the Arduino board such that user can perform inference locally with low latency.

\subsection{Conclusion}
In conclusion, this project demonstrate that SFL shows clear advantage over FL because it can implement a larger CNN model. It can achieve both great utility than FL and better preserve user data privacy compared to centralized training.

\subsection{Limitations and Future work}

\textbf{Limitations}
There is a workflow we find online for the model transformation between different framework: Pytorch -> ONNX -> TensorFlow -> TensorFlowlite (TFlite). This transformation is not suitable for final deployment on a micro-controller as the \emph{Transpose} operation is not yet supported by TFlite micro library, which is needed to handle the different layout of the shape of weight in conv2d layer in Pytorch and TensorFlow.
Thus, we can only take this approach: Pytorch -> Keras -> TFlite. Convert Pytorch model into Keras needs reinstantiate of the model in Keras and load the Pytorch weight in.
However, as the Keras and Pytorch have different weight format, we spent extra effort in the conversion. Because of this delay, we don't have time for the final deployment \& on-board testing.

\textbf{Future works}
\begin{itemize}
\item Bluetooth Communication Support for more convenient use
\item Training Data Labeling and Data Storage Support. Get user’s raw data stored on the edge device and labeled at the same time for SFL training.
\item Train multi clients in parallel. (Currently only support sequential training)
\item Instant deployment. Once global model converges, the server generates deployable model (compressed/quantized) and send it to clients.
\end{itemize}



\bibliographystyle{ACM-Reference-Format}
\bibliography{sample-base}










\end{document}